\documentclass[letterpaper]{article}
\usepackage{aaai2027}
\usepackage[hyphens]{url}
\usepackage{graphicx}
\usepackage{adjustbox}
\usepackage{natbib}
\usepackage[utf8]{inputenc}
\usepackage[T1]{fontenc}
\usepackage{booktabs}
\usepackage{amsmath}
\usepackage{amsfonts}
\usepackage{amssymb}
\usepackage{mathrsfs}
\usepackage{microtype}

\newcommand{\dpsnr}{\Delta\mathrm{PSNR}}
\newcommand{\doff}{\Delta\mathrm{OffRel}}
\newcommand{\offrel}{\mathrm{OffRel}}
\newcommand{\diag}{\operatorname{diag}}
\newcommand{\maybeincludegraphics}[3]{%
\IfFileExists{#1}{\includegraphics[width=#2\linewidth]{#1}}{\fbox{\parbox{0.90\linewidth}{\centering #3}}}%
}

\nocopyright
\title{Latent-Frequency Validity: Fast Spectral Editing with Screened Video-VAE Transfer Operators}

\author{
Bowen Xue\textsuperscript{\rm 1,2},
Jiafeng Xiong\textsuperscript{\rm 1},
Xin Quan\textsuperscript{\rm 1,3}
}

\affiliations{
\textsuperscript{\rm 1}University of Manchester, United Kingdom\\
\textsuperscript{\rm 2}NVIDIA, United States\\
\textsuperscript{\rm 3}Idiap Research Institute, Switzerland\\
}

\begin{document}
\maketitle

\begin{abstract}
Direct spectral editing in video-VAE latents can control noise, flicker, smoothness, and frequency content without a decode--filter--reencode pass. However, video VAEs may redistribute pixel-space frequency bands across latent channels, and latent edits can disrupt VAE round-trip dynamics. We introduce \emph{latent-frequency validity} (LFV), which learns a compact VAE-specific spectral response and deploys it only when it improves decoded-target fidelity without worsening round-trip drift. LFV follows a validation-selected path from a diagonal per-frequency calibrator (C1) to full channel mixing (CM), making cross-channel capacity a controllable per-edit resource. Across 544 VAE--edit cells spanning six spectral families, LFV emits 423 cheap operators: 277 are handled by C1, while 146 (34.5\% of emitted operators) require channel mixing. On the primary 120-cell radial sweep, 99/100 emitted operators pass source-video-grouped held-out evaluation. Across five additional filter families, all 323 emitted operators pass held-out evaluation. Fully frozen OpenVid-fitted operators, including the validation-selected path coefficient, pass all 20 tested CogVideoX and HunyuanVideo generated-domain cells without adaptation. The selected response matches direct latent-filter latency and is about $3\times$ faster than pixel filter--reencode. The resulting maps reveal distinct VAE regimes, including strongly channel-coupled CogVideoX responses and a sharp Open-Sora high-band stability frontier.
\end{abstract}

\section{Introduction}
Spectral operations are useful primitives for video generation and reconstruction. They can attenuate coarse noise, suppress flicker, smooth a decoded sequence, or isolate selected temporal and spatial bands. With a fixed pretrained video VAE, the reference implementation is straightforward: decode the latent video, apply the desired pixel-space filter, and re-encode the result. Applying a Fourier mask directly to the latent tensor is far cheaper and fits naturally inside a latent generation pipeline.

Yet the two implementations need not agree. A video VAE learns its own spatiotemporal analysis and synthesis transform, so a pixel-space frequency band can shift across latent frequencies, become attenuated, or mix across channels. This creates a model-specific transfer problem: which spectral edits admit an efficient latent realization, what response is required, and where should the system use the reference path instead? We capture this structure with a \emph{VAE-specific operating map}.

We formulate the operating map through \emph{latent-frequency validity} (LFV). For each VAE and pixel-space filter, decode--filter--reencode defines the target operation. LFV evaluates a learned latent response using two paired quantities: decoded-target fidelity and decode--encode round-trip drift. The two criteria are complementary. A more expressive operator may approximate the target better while moving the latent into a less stable region, so target fidelity alone is insufficient for deployment.

To model the VAE response, C1 fits an independent gain for each latent channel and frequency, testing whether the transfer is channel-separable. Full channel mixing captures cross-channel response that C1 cannot represent. We connect these endpoints with a C1--CM path and let validation select the amount of channel-mixing capacity supported by each VAE--edit cell. The path is compact and interpretable: $\alpha=0$ gives the diagonal response, $\alpha=1$ gives full mixing, and intermediate values damp cross-channel residuals when the unrestricted estimator is unnecessarily aggressive.

The resulting map has three outcomes. \emph{C1-sufficient} cells require only per-channel calibration. \emph{Path-rescued} cells require cross-channel response and become deployable through the C1--CM path. \emph{Fallback} cells are routed to the reference branch because the tested cheap family does not satisfy both LFV criteria. Across 544 cells, 277 are C1-sufficient, 146 are path-rescued, and 121 fall back; therefore, 34.5\% of emitted operators rely on channel mixing. The distribution is highly structured: CogVideoX is predominantly channel-coupled, WAN is largely diagonal with several damping-sensitive cases, HunyuanVideo combines both regimes, and Open-Sora forms a sharp high-frequency round-trip frontier.

\paragraph{Contributions.}
We make three contributions. First, LFV provides a paired, statistically tested criterion for learning and deploying fast spectral operators in fixed video-VAE latents, together with a source-coordinate stride-aware reference. Second, we introduce a C1--channel-mixing path that turns cross-channel capacity into a per-edit control variable and recovers operators that a diagonal response cannot express. Third, we present a 544-cell study across four video VAEs and six spectral families, including source-video-grouped evaluation, fully frozen transfer to generated inputs, repeated-cycle and perceptual analysis, and four-VAE runtime profiling. The resulting operating maps characterize both the capabilities and the latent spectral geometry of modern video VAEs.

\section{Related Work}
\paragraph{Video VAEs and latent generation.}
VAEs and latent diffusion separate perceptual compression from generative modeling \citep{kingma2014vae,rombach2022ldm}, and video systems extend this idea with spatiotemporal autoencoders \citep{blattmann2023align,zhao2024cvvae}. CogVideoX, Open-Sora, HunyuanVideo, and Wan use distinct video VAEs \citep{yang2025cogvideox,zheng2024opensora,kong2024hunyuanvideo,wan2025}. We freeze these pretrained VAEs and ask whether post-hoc latent spectral operators can replace pixel filtering and re-encoding.

\paragraph{Spectral structure of autoencoder latents.}
Frequency-aware autoencoders improve reconstruction or diffusability through explicit frequency paths, equivariance, and spectrum shaping \citep{lin2023frequencyvae,kouzelis2025eqvae,skorokhodov2025diffusability}. Other analyses identify asymmetric high-frequency encoding and decoding or regularize spatiotemporal latent spectra \citep{lai2026freqperspective,li2025wfvae,liu2026ssvae,ning2026spectrummatching}. These methods modify training or representation learning. LFV instead keeps the VAE fixed and measures whether a post-hoc latent operator realizes a specified decoded pixel operation.

\paragraph{Frequency-domain operators and interventions.}
Fourier operators range from elementwise filtering to per-frequency channel mixing \citep{li2021fno,rao2021gfnet,guibas2022afno}. Frequency cues have also been used for video initialization, inversion, denoising, and editing trajectories \citep{wu2024freeinit,koo2024flexiedit,ren2025fds,huang2025fresca,zhu2025fade}. Our C1 and full-CM endpoints instead estimate the VAE-specific latent response induced by a fixed pixel-space filter, and validation selects the response used at deployment.

\paragraph{Decode consistency and evaluation.}
Nonlinear VAE decoders can make latent operations inconsistent with pixel-space counterparts \citep{bradbury2026pelc}, while repeated pixel--latent transitions may accumulate reconstruction or frequency drift \citep{almog2025reedvae,liao2026freqedit,wang2026ditdrift}. LPIPS, FVD, and VBench assess perceptual, distributional, or multidimensional video quality \citep{zhang2018lpips,unterthiner2018fvd,huang2024vbench}. Because our task is paired approximation of a fixed operator on the same clips, decoded-target PSNR and relative round-trip drift define the deployment gates; perceptual and repeated-cycle measures serve as corroborating diagnostics.

\section{Latent-Frequency Validity}
Let the encoder and decoder of a fixed video VAE be
\begin{equation}
\begin{aligned}
E &: \mathbb{R}^{T\times H\times W\times 3}\rightarrow\mathbb{R}^{C\times T'\times H'\times W'},\\
D &: \mathbb{R}^{C\times T'\times H'\times W'}\rightarrow\mathbb{R}^{T\times H\times W\times 3}.
\end{aligned}
\end{equation}
For a pixel-space spectral filter $\Phi_\theta$, the target latent for clip $x$ is
\begin{equation}
\label{eq:target}
z_\theta^\star(x)=E\!\left(\Phi_\theta(D(E(x)))\right).
\end{equation}
This decode--filter--reencode path defines the spectral operation in the VAE's own latent space. A source-filter-reencode target $E(\Phi_\theta(x))$ provides an independent target-sensitivity check. Let $\mathscr{F}$ be the 3D FFT over temporal and spatial latent axes. Our primary cheap reference is the same-coordinate latent filter
\begin{equation}
\label{eq:lf}
\mathcal{L}^{\mathrm{LF}}_\theta(z)=\mathscr{F}^{-1}\!\left(M_\theta\odot\mathscr{F}(z)\right),
\end{equation}
which applies the same numerical mask parameterization directly in latent FFT coordinates. This is the natural drop-in shortcut used by a latent pipeline. To separate direct coordinate reuse from source-video frequency alignment, we also construct a stride-aware reference. If the VAE compression strides are $(s_t,s_h,s_w)$ and a latent FFT bin has coordinates $(f_t,f_h,f_w)$ in cycles per latent sample, its source-video coordinates are
\begin{equation}
\label{eq:source_freq}
\mathbf{f}^{x}=\left(f_t/s_t,\,f_h/s_h,\,f_w/s_w\right),
\qquad r_x=\|\mathbf{f}^{x}\|_2.
\end{equation}
The \emph{stride-aware latent filter} applies the pixel-space mask as $M_\theta(r_x)$ on latent FFT bins. Thus a reported latent-grid center is not interpreted as a universal cycles-per-pixel value: its source-video support is VAE-specific and follows Eq.~\eqref{eq:source_freq}. The primary map measures improvement over the direct shortcut in Eq.~\eqref{eq:lf}, while the stride-aware comparison isolates gains that remain after correcting this sampling geometry. For an edited latent $\hat z$, round-trip drift is
\begin{equation}
\label{eq:offrel}
\offrel(\hat z)=\frac{\|\hat z-E(D(\hat z))\|_2}{\max(\|\hat z\|_2,\varepsilon)}.
\end{equation}
We compare every candidate with the same-coordinate latent filter on the same clip:
\begin{equation}
\label{eq:deltas}
\begin{aligned}
\dpsnr &= \mathrm{PSNR}_{\mathrm{method}}-\mathrm{PSNR}_{\mathrm{LF}},\\
\doff &= \offrel_{\mathrm{method}}-\offrel_{\mathrm{LF}}.
\end{aligned}
\end{equation}
Paired bootstrap intervals yield the lower 95\% bound $L^{\mathrm{psnr}}$ and upper 95\% bound $U^{\mathrm{off}}$. A candidate passes LFV when
\begin{equation}
\label{eq:gate}
L^{\mathrm{psnr}}>0,\qquad U^{\mathrm{off}}\le 0.
\end{equation}
Equation~\eqref{eq:gate} defines a strict win over the latent shortcut: the learned response must improve target fidelity and preserve or improve round-trip stability. The same criterion is recomputed against the source-coordinate reference in the stride-aware comparison.

\section{Screened C1--Channel Mixing}
For fitting pairs $(z_b,z_{\theta,b}^{\star})$, write $F_{b,c,\omega}=\mathscr{F}(z_b)_{c,\omega}$ and $G_{b,c,\omega}=\mathscr{F}(z_{\theta,b}^{\star})_{c,\omega}$. C1 fits one complex gain per channel and frequency:
\begin{equation}
\label{eq:c1_obj}
\min_{a_{c,\omega}}\sum_b|a_{c,\omega}F_{b,c,\omega}-G_{b,c,\omega}|^2+\lambda_{\mathrm{C1}}|a_{c,\omega}-a^0_{c,\omega}|^2,
\end{equation}
with solution
\begin{equation}
\label{eq:c1_closed}
a_{c,\omega}^{\star}=\frac{\sum_b\overline{F_{b,c,\omega}}G_{b,c,\omega}+\lambda_{\mathrm{C1}} a^0_{c,\omega}}{\sum_b|F_{b,c,\omega}|^2+\lambda_{\mathrm{C1}}}.
\end{equation}
Let $D_{\theta,\omega}=\diag(a_{1,\omega}^{\star},\ldots,a_{C,\omega}^{\star})$.

Full channel mixing fits a matrix at each frequency. With $F_\omega=[f_{1,\omega},\ldots,f_{B,\omega}]$ and $G_\omega=[g_{1,\omega},\ldots,g_{B,\omega}]$,
\begin{equation}
\label{eq:cm_obj}
\min_{K_\omega}\|K_\omega F_\omega-G_\omega\|_F^2+\lambda_{\mathrm{CM}}\|K_\omega-K^0_\omega\|_F^2,
\end{equation}
whose ridge solution is
\begin{equation}
\label{eq:cm_closed}
K_\omega^{\mathrm{CM}}=(G_\omega F_\omega^\ast+\lambda_{\mathrm{CM}} K^0_\omega)(F_\omega F_\omega^\ast+\lambda_{\mathrm{CM}} I)^{-1}.
\end{equation}
With matched priors and ridge weights, imposing a diagonal constraint on Eq.~\eqref{eq:cm_obj} reduces it to independent C1 solves. Our final endpoints use separately selected regularization, so we treat them as complementary fitted hypotheses: C1 tests channel-separable response, while full CM estimates complete cross-channel transfer. The matched-regularization audit in Table~\ref{tab:stress_audits} preserves the same coverage conclusion. Positive ridge weights make both systems uniquely solvable. A residual view makes the capacity difference explicit. If the target response at one frequency is $g=Kf+\xi$, with $\mathbb{E}[\xi f^\ast]=0$ and source covariance $\Sigma_f$, then any diagonal response $R$ incurs
\begin{equation}
\label{eq:residual}
\mathbb{E}\|Rf-g\|_2^2=\operatorname{tr}\!\left((R-K)\Sigma_f(R-K)^\ast\right)+\mathbb{E}\|\xi\|_2^2.
\end{equation}
Off-diagonal target response therefore creates an approximation residual that C1 cannot remove. Channel mixing directly addresses this residual, while LFV determines how much of the fitted response should be used for deployment.

Our candidate family is
\begin{equation}
\label{eq:path}
K_\omega(\alpha)=(1-\alpha)D_{\theta,\omega}+\alpha K_\omega^{\mathrm{CM}},\qquad \alpha\in\mathcal{A}.
\end{equation}
Endpoints are fitted on the fitting split. Validation evaluates the capacity path, retains alphas satisfying Eq.~\eqref{eq:gate}, and chooses the candidate with the largest fidelity lower bound, breaking ties by the smaller safety upper bound. If the path does not contain an eligible response, the cell routes to the reference branch. Test clips only evaluate the frozen choice. The selected matrix response is Hermitian-projected before inverse FFT so edited latents remain real.

The three stages have distinct roles: fitting estimates the VAE response, validation chooses the supported amount of channel-mixing capacity, and test measures the frozen operator. At inference, deployment reduces to one fixed frequency response; there is no per-input search over $\alpha$.

\section{Experimental Design}
\paragraph{Models, data, and nested splits.}
We evaluate WAN, CogVideoX, Open-Sora v1.3, and HunyuanVideo on OpenVid1M \citep{nan2025openvid}. Endpoint fitting uses 256 clips, validation selection uses 128, and final reporting uses 128 untouched clips. All comparisons remain paired within clip. Our primary confidence intervals use a source-video group bootstrap: source IDs are resampled with replacement, all clips from each sampled source are retained, and method differences remain paired within each clip. Paired clip bootstrap is reported as a sensitivity analysis. A cell is \emph{emitted} only when validation selects an LFV-eligible path point; it is \emph{path-rescued} when C1 fails but the selected path passes, and \emph{fallback} when no validation-eligible point exists.

\paragraph{Primary and breadth frequency settings.}
The primary sweep contains 30 radial band-pass centers per VAE, $0.025,0.050,\ldots,0.750$, for 120 cells. The canonical grid uses raw \texttt{torch.fft.fftfreq} cycles per latent sample and radial coordinate $r=\sqrt{f_t^2+f_h^2+f_w^2}$, so the joint spatiotemporal corner exceeds the single-axis Nyquist value. Equation~\eqref{eq:source_freq} gives the corresponding source-video interpretation used by the stride-aware reference. Bands use width 0.10 and cosine transition width 0.02. Breadth experiments add 424 cells: 60 radial low-pass, 60 radial high-pass, 120 notch, 108 spatial-band, and 76 temporal-band cells.

\paragraph{Mechanism and generalization studies.}
A five-center slice $\{0.05,0.15,0.25,0.35,0.45\}$ across four VAEs provides a compact 20-cell mechanism study of C1 retention, CM-only upgrades, and stability frontiers. Additional evaluations regroup clips by source identity, vary band width, clip length, resolution, and sampling stride, stratify by motion, texture, and compression proxies, and include phase-based stress tests. Each evaluation preserves the fitting--validation--test separation.

\paragraph{Frozen generated-domain transfer.}
We test whether the learned operator bank transfers from reconstruction clips to generated inputs without any domain-specific adaptation. For each evaluated cell, we freeze the C1 and CM endpoints and the validation-selected $\alpha$ from the OpenVid study. We then evaluate this fixed operator on 64 generated samples per cell, with no refitting and no reselection. Two protocols are considered: direct generated latents and generated videos re-encoded by the corresponding VAE. We evaluate five centers for each of CogVideoX and HunyuanVideo under both protocols, for 20 generated-domain cells in total.

\paragraph{Decoded and temporal diagnostics.}
We evaluate 18 representative emitted cells on the full 128-clip test split using LPIPS, temporal-difference error, temporal high-frequency error, and five-cycle VAE drift. These measurements provide decoded and temporal evidence for the selected responses without participating in operator selection.

\paragraph{Hyperparameters and runtime.}
Inputs contain 16 frames at $256\times256$ and produce latents of shape $16\times4\times32\times32$. The alpha grid is $\{0,0.1,0.2,0.35,0.5,0.7,0.85,1\}$. C1 uses ridge $10^{-4}$ with zero prior; full CM uses ridge $2\times10^{-2}$, identity prior, real gain mode, spectral-norm clamp 1.5, and Hermitian projection. LFV intervals use 4000 paired bootstrap replicates. Runtime is measured at center 0.25 on an NVIDIA L40S in FP16, batch size 1, with 120 timed runs per method and VAE. All \texttt{from\_z} branches begin from an existing latent and include the spectral operation followed by one output decode. Diffusion sampling, data loading, and offline endpoint fitting are excluded. Pixel filter--reencode measures decode $\rightarrow$ pixel filter $\rightarrow$ encode $\rightarrow$ output decode; LF projection measures latent filter $\rightarrow$ decode $\rightarrow$ encode $\rightarrow$ output decode.

\section{Results}
\subsection{LFV Reveals Three VAE-Specific Regimes}
\begin{figure*}[t]
\centering
\maybeincludegraphics{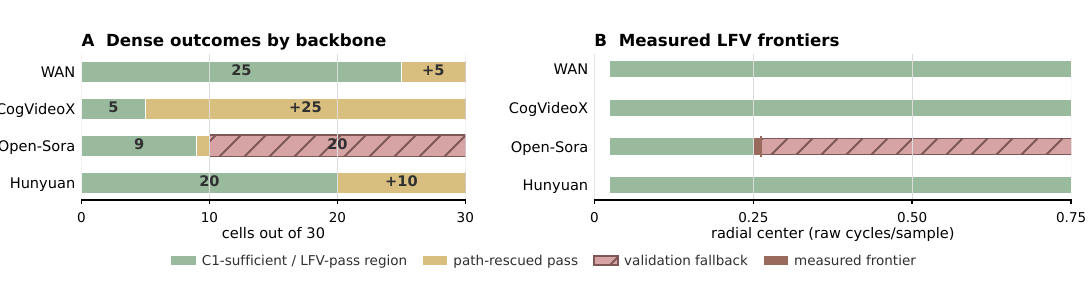}{0.98}{Dense LFV operating-map figure. Include \texttt{lfv\_figure1\_all\_backbones\_wideshort\_v21.pdf} with the source package.}
\caption{Dense LFV map. C1-sufficient and path-rescued cells form the emitted region; Open-Sora v1.3 exhibits a refined safety frontier at $(0.25,0.2625]$, whereas no fallback is observed for the other three VAEs through center 0.75.}
\label{fig:map}
\end{figure*}

Table~\ref{tab:coverage} gives the primary operating map. C1 alone supports 59/120 cells; full channel mixing expands this to 86/120. The validation-selected path reaches 100 cells, of which 99 pass the held-out source-video-grouped LFV test. Paired clip bootstrap gives 100/100, and the two analyses agree on 119/120 map decisions; the sole change is the thin WAN 0.525 boundary cell. Most importantly, channel mixing creates 41 deployable operators beyond the diagonal endpoint. Figure~\ref{fig:map} shows how these gains organize into distinct VAE regimes rather than a uniform global response.

\begin{table}[t]
\centering
\caption{Primary 120-cell radial map. Group pass is the main statistic and resamples source-video IDs; clip pass is a paired clip-bootstrap sensitivity analysis.}
\label{tab:coverage}
\small
\begin{adjustbox}{max width=\linewidth}
\begin{tabular}{lrrrrrrr}
\toprule
Model & C1 & Full CM & Emitted & Group pass & Clip pass & Rescued & Fallback \\
\midrule
WAN & 25 & 18 & 30 & 29/30 & 30/30 & 5 & 0 \\
CogVideoX & 5 & 30 & 30 & 30/30 & 30/30 & 25 & 0 \\
Open-Sora v1.3 & 9 & 10 & 10 & 10/10 & 10/10 & 1 & 20 \\
HunyuanVideo & 20 & 28 & 30 & 30/30 & 30/30 & 10 & 0 \\
\midrule
Total & 59 & 86 & \textbf{100} & \textbf{99/100} & \textbf{100/100} & \textbf{41} & \textbf{20} \\
\bottomrule
\end{tabular}
\end{adjustbox}
\end{table}

The 20-cell mechanism slice makes the tradeoff explicit (Table~\ref{tab:audit}). Full CM recovers all five upgrades unavailable to C1, but loses two WAN cells that C1 passes. The path retains all 13 C1 passes and all five CM-only upgrades, reaching 18/20. Anchored, off-diagonal-shrunk, and richer covered variants also reach 18/20 but do not improve coverage; a dense 120-cell rerun likewise adds no cells beyond the convex path. Table~\ref{tab:stress_audits} tests the primary result under source dependence, source-coordinate frequency alignment, matched regularization, and a richer operator family.

\begin{table}[t]
\centering
\caption{Mechanism study on five centers for each VAE.}
\label{tab:audit}
\small
\begin{adjustbox}{max width=\linewidth}
\begin{tabular}{lrrrr}
\toprule
Method & Pass & C1 retained & CM-only retained & Fallback \\
\midrule
C1 & 13/20 & 13/13 & 0/5 & 7 \\
Full CM & 16/20 & 11/13 & 5/5 & 4 \\
C1--CM path & \textbf{18/20} & \textbf{13/13} & \textbf{5/5} & \textbf{2} \\
Richer covered family & 18/20 & 13/13 & 5/5 & 2 \\
\bottomrule
\end{tabular}
\end{adjustbox}
\end{table}

\begin{table}[t]
\centering
\caption{Primary source-grouped reporting and complementary controls. The stride-aware row compares the frozen selected path with the best validation-selected source-coordinate latent filter.}
\label{tab:stress_audits}
\small
\begin{adjustbox}{max width=\linewidth}
\begin{tabular}{lcc}
\toprule
Evaluation & Result & Main observation \\
\midrule
Source-video group bootstrap & 99/100 & operating map is preserved \\
Stride-aware latent reference & 83/100 & learned response exceeds coordinate alignment \\
Matched regularization & 100/100 & full CM remains 86/120 \\
Richer covered family & 100/100 & no additional expansions \\
\bottomrule
\end{tabular}
\end{adjustbox}
\end{table}

Source-video grouping is the main statistic and retains 99/100 emitted decisions. The complementary controls preserve the central conclusion: matching endpoint regularization yields 100/100 path coverage while full CM remains at 86/120, and richer covered families add no cells. Most notably, 83/100 emitted decisions retain both LFV advantages over the stride-aware source-coordinate filter. The learned response therefore captures VAE-specific spectral transfer beyond a simple correction for temporal and spatial stride.

\paragraph{Validation-to-test transfer.}
Table~\ref{tab:val_audit} shows that validation margins carry cleanly to held-out data. Ordinary passes, thin passes, and channel-mixing rescues retain their decisions after $\alpha$ is frozen, while Open-Sora 0.35 remains outside the selected path.

\begin{table*}[t]
\centering
\caption{Representative paired clip-bootstrap validation-to-test margins. Each selected $\alpha$ is chosen on validation and evaluated once on the untouched test split.}
\label{tab:val_audit}
\small
\begin{adjustbox}{max width=\linewidth}
\begin{tabular}{lrrrrrl}
\toprule
Cell & $\alpha$ & Val PSNR LB & Val OffRel UB & Test PSNR LB & Test OffRel UB & Outcome \\
\midrule
WAN 0.05 & 0.20 & +3.488 & -0.305 & +3.626 & -0.319 & pass \\
WAN 0.25 & 0.50 & +2.661 & -0.047 & +2.757 & -0.050 & thin pass \\
CogVideoX 0.15 & 1.00 & +6.792 & -4.528 & +6.754 & -4.521 & rescued \\
Open-Sora 0.25 & 1.00 & +5.704 & -0.112 & +5.372 & -0.124 & rescued \\
HunyuanVideo 0.25 & 1.00 & +5.510 & -0.003 & +5.408 & -0.030 & thin pass \\
\bottomrule
\end{tabular}
\end{adjustbox}
\end{table*}

\subsection{Channel Mixing Works Best as a Controlled Capacity Dial}
Full channel mixing supplies the capacity needed by CogVideoX and many spatial or temporal filters, but the best response is not always the endpoint $\alpha=1$. Table~\ref{tab:wan_counter} shows two WAN cells in which the full-CM fit overuses cross-channel residuals. Intermediate path points improve fidelity over the latent shortcut while recovering the stable behavior of the diagonal response. The path therefore acts as a data-driven capacity dial, not merely a choice between two discrete estimators.
\begin{table*}[t]
\centering
\caption{WAN counterexamples showing why fitted full CM cannot simply replace C1. Intervals are held-out paired 95\% confidence intervals.}
\label{tab:wan_counter}
\small
\begin{adjustbox}{max width=\linewidth}
\begin{tabular}{llccl}
\toprule
Cell & Method & $\dpsnr$ mean [95\% CI] & $\doff$ mean [95\% CI] & Decision \\
\midrule
WAN 0.05 & C1 & $3.954,[3.457,4.450]$ & $-0.413,[-0.448,-0.380]$ & pass \\
WAN 0.05 & full CM & $2.043,[1.927,2.167]$ & $+0.079,[0.047,0.107]$ & safety-limited \\
WAN 0.05 & selected path & $4.078,[3.616,4.540]$ & $-0.351,[-0.386,-0.319]$ & pass \\
\midrule
WAN 0.25 & C1 & $2.908,[2.457,3.355]$ & $-0.218,[-0.242,-0.195]$ & pass \\
WAN 0.25 & full CM & $2.600,[2.509,2.695]$ & $+0.095,[0.074,0.117]$ & safety-limited \\
WAN 0.25 & selected path & $3.056,[2.754,3.370]$ & $-0.075,[-0.099,-0.051]$ & thin pass \\
\bottomrule
\end{tabular}
\end{adjustbox}
\end{table*}

\paragraph{Residual-control diagnostics.}
Off-diagonal shrinkage makes this interpretation concrete. In WAN 0.05, reducing the off-diagonal ratio from $0.853$ to $0.372$ improves both $\dpsnr$ ($+3.568$ to $+8.334$) and $\doff$ ($-0.124$ to $-0.408$), showing that damping removes harmful residual mixing. CogVideoX 0.15 remains strongly improved even under heavy shrinkage ($\dpsnr\ge+6.89$, $\doff<-4.70$), revealing a genuinely cross-channel target response. Open-Sora 0.35 and 0.45 remain beyond the stability boundary at every tested ratio, completing the three-regime picture.

\subsection{Cross-Channel Response Generalizes Across Spectral Operators}
Having established how the path controls channel-mixing capacity on the primary map, we next test whether the same response structure appears for other spectral operators.
The learned response extends well beyond radial band-pass filters (Table~\ref{tab:families}). Across 424 additional cells, LFV emits 323 cheap operators, and all 323 pass the held-out LFV gate. Of these, 218 are diagonal-sufficient and 105 are created by the channel-mixing path. Cross-channel response is especially valuable for spatial and temporal bands, which contribute 38 and 43 rescues, respectively; high-pass filters add another 23. These results show that channel coupling is a recurring property of video-VAE spectral transfer rather than an artifact of one mask family.

\begin{table}[t]
\centering
\caption{Held-out LFV outcomes on 424 additional spectral-edit cells. Each emitted operator is selected on validation and evaluated once on the held-out split.}
\label{tab:families}
\small
\begin{adjustbox}{max width=\linewidth}
\begin{tabular}{lrrrrrr}
\toprule
Family & Cells & C1 sufficient & Rescued & Emitted & Held-out pass & Fallback \\
\midrule
Radial low-pass & 60 & 59 & 0 & 59 & 59/59 & 1 \\
Radial high-pass & 60 & 18 & 23 & 41 & 41/41 & 19 \\
Radial notch & 120 & 71 & 1 & 72 & 72/72 & 48 \\
Spatial band & 108 & 49 & 38 & 87 & 87/87 & 21 \\
Temporal band & 76 & 21 & 43 & 64 & 64/64 & 12 \\
\midrule
Total & \textbf{424} & \textbf{218} & \textbf{105} & \textbf{323} & \textbf{323/323} & \textbf{101} \\
\bottomrule
\end{tabular}
\end{adjustbox}
\end{table}

The 105 additional rescues are distributed across both models and operators: CogVideoX contributes 55, HunyuanVideo 23, Open-Sora 16, and WAN 11. Low-pass edits are predominantly diagonal-sufficient, whereas spatial and temporal bands expose the largest need for cross-channel transfer. High-pass and notch families produce sharper boundaries, particularly for Open-Sora. The operator family therefore reveals a structured interaction between the VAE architecture and the geometry of the requested filter.

\paragraph{A unified 544-cell view.}
Combining the primary and breadth blocks gives the clearest summary of the method (Table~\ref{tab:three_regimes}). Of 544 tested VAE--edit cells, 277 are handled by diagonal C1 and 146 are unlocked by channel mixing, yielding 423 clip-level held-out LFV passes. More than one third of these operators ($146/423=34.5\%$) would be unavailable to a diagonal method. The C1--CM path is therefore a substantive capacity expansion: it converts cross-channel spectral structure into deployable edits while routing the remaining 121 cells to the reference branch.

\begin{table}[t]
\centering
\caption{The three LFV regimes across all reported frequency-family cells.}
\label{tab:three_regimes}
\small
\begin{adjustbox}{max width=\linewidth}
\begin{tabular}{lrrrrr}
\toprule
Block & Cells & C1 sufficient & Rescued & Emitted & Fallback \\
\midrule
Primary radial band-pass & 120 & 59 & 41 & 100 & 20 \\
Additional families & 424 & 218 & 105 & 323 & 101 \\
\midrule
All tested cells & \textbf{544} & \textbf{277} & \textbf{146} & \textbf{423} & \textbf{121} \\
\bottomrule
\end{tabular}
\end{adjustbox}
\end{table}

\subsection{The Map Reveals Video-VAE Spectral Geometry}
The aggregate map exposes two complementary axes of structure. Along the \emph{filter} axis, low-pass edits are almost entirely diagonal, whereas spatial and temporal bands frequently require cross-channel response. Along the \emph{model} axis, CogVideoX contributes 80 of the 146 rescues across the primary and breadth blocks, making it the clearest channel-coupled VAE. WAN contributes only 16 rescues and is mostly diagonal-sufficient; its characteristic behavior is not missing capacity but the need to damp an unnecessarily aggressive full-CM fit. HunyuanVideo occupies a mixed regime with 33 rescues and broad coverage. Open-Sora contributes 17 rescues but concentrates the high-frequency stability boundaries.

These patterns are consistent with a learned spectral transfer geometry rather than a scalar frequency remapping. The fitted transfer-map diagonal mass has a moderate association with C1 fidelity (Spearman $\rho=0.501$ over 20 mechanism cells), while the residual analysis in Eq.~\eqref{eq:residual} explains why cross-channel terms can reduce approximation error. The operating map therefore provides more than a collection of pass counts: it is a compact diagnostic of how each VAE redistributes temporal and spatial bands across latent channels. This diagnostic view explains why a single universal mask is insufficient and why a small, VAE-specific response bank is effective.

\subsection{Frozen Operators Transfer to Generated Inputs}
The operating map is learned on OpenVid reconstruction clips, but the resulting responses also apply to generated inputs without adaptation. We freeze the C1 and CM endpoints and the validation-selected $\alpha$ from the OpenVid study, then evaluate the same operator on 64 generated samples per cell. No generated sample is used for fitting or selection.

Table~\ref{tab:generated} reports two complementary input routes. In \emph{generated-latent direct}, the frozen response is applied to saved diffusion latents. In \emph{generated-video source}, generated videos are encoded by the corresponding VAE before applying the same frozen response. CogVideoX and HunyuanVideo pass all five tested centers under both routes, for 20/20 held-out generated-domain passes. The result shows that the learned spectral transfer is not confined to reconstruction latents from the fitting distribution; it remains effective on latents and videos produced by the corresponding generative models.

\begin{table}[t]
\centering
\caption{Fully frozen generated-domain transfer. C1 and CM endpoints and the OpenVid-selected $\alpha$ are evaluated on 64 generated samples per cell without refitting or reselection.}
\label{tab:generated}
\small
\begin{adjustbox}{max width=\linewidth}
\begin{tabular}{lrrr}
\toprule
Model & Direct generated latent & Generated video & Total \\
\midrule
CogVideoX & 5/5 & 5/5 & 10/10 \\
HunyuanVideo & 5/5 & 5/5 & 10/10 \\
\midrule
Total & \textbf{10/10} & \textbf{10/10} & \textbf{20/20} \\
\bottomrule
\end{tabular}
\end{adjustbox}
\end{table}

\subsection{LFV Discovers a Sharp Open-Sora High-Band Frontier}
The same map that identifies deployable operators also localizes where the fast path should end.
Open-Sora reveals the sharpest boundary in the operating map. The selected operator passes at 0.25, while half-step refinement places the transition in $(0.25,0.2625]$ (Table~\ref{tab:frontier}). Beyond the boundary, target-fidelity lower bounds remain strongly positive but round-trip upper bounds cross zero. LFV therefore separates a region where additional channel-mixing capacity is useful from a region where the cheap latent path should hand off to the reference implementation.

\begin{table}[t]
\centering
\caption{Open-Sora half-step refinement. Every tested center above 0.25 is safety-limited despite positive fidelity lower bounds.}
\label{tab:frontier}
\small
\begin{adjustbox}{max width=\linewidth}
\begin{tabular}{rrrrl}
\toprule
Center & $\alpha$ & PSNR LB & OffRel UB & Decision \\
\midrule
0.2500 & 1.0 & +5.707 & -0.112 & pass \\
0.2625 & -- & +5.569 & +0.025 & fallback \\
0.2750 & -- & +5.394 & +0.131 & fallback \\
0.2875 & -- & +5.812 & +0.285 & fallback \\
0.3000 & -- & +5.506 & +0.409 & fallback \\
0.3125 & -- & +4.809 & +0.448 & fallback \\
0.3250 & -- & +4.562 & +0.487 & fallback \\
0.3375 & -- & +4.178 & +0.533 & fallback \\
0.3500 & -- & +3.731 & +0.506 & fallback \\
0.3625 & -- & +3.704 & +0.563 & fallback \\
\bottomrule
\end{tabular}
\end{adjustbox}
\end{table}
The rejected high-band edit is not numerically negligible: mean relative edit energy is approximately $0.976$ at 0.25, $0.975$ at 0.275, $0.994$ at 0.525, and $0.9999$ at 0.75.

The decision also replicates under source-video grouping. Across the complete dense map, regrouping by source identity preserves 119/120 decisions; the sole change is WAN 0.525, whose grouped test safety upper bound is $+0.0027$. An independently regenerated Open-Sora boundary split refits endpoints and reselects $\alpha$ on validation, yet reproduces the pass at 0.25 and safety fallbacks above it (Table~\ref{tab:sourceid}).

\begin{table}[t]
\centering
\caption{Source-ID-grouped replication of the Open-Sora boundary neighborhood.}
\label{tab:sourceid}
\small
\begin{tabular}{rrrl}
\toprule
Center & Val PSNR LB & Val OffRel UB & Decision \\
\midrule
0.2500 & +5.298 & -0.103 & pass \\
0.2625 & +5.190 & +0.042 & fallback \\
0.2750 & +5.062 & +0.159 & fallback \\
0.3000 & +5.071 & +0.439 & fallback \\
\bottomrule
\end{tabular}
\end{table}

The same frontier structure appears under broader settings. Width and input variations yield 48/72 and 45/50 passes, with the boundary again dominated by round-trip behavior. Motion subgroup decisions agree with the main map in 18/20 cells, while texture and compression groupings agree in 17/20. Six of eight phase-based stress tests also land outside the cheap operating region.

\subsection{OffRel Predicts Repeated Drift and Matches Decoded Quality}
One-round OffRel provides an efficient proxy for repeated VAE behavior. Across 9984 method--clip observations, its Spearman correlation with five-cycle drift is $\rho=0.948$ with 95\% CI $[0.945,0.950]$; within-family correlations remain between $0.918$ and $0.947$. This strong agreement makes the one-cycle statistic practical for dense validation while preserving the behavior of a more expensive multi-cycle diagnostic.

At rejected Open-Sora center 0.35, full CM improves PSNR and LPIPS over the latent filter (24.935 versus 21.125 and 0.782 versus 0.939), but OffRel and five-cycle drift remain worse (1.465 versus 0.974 and 2.652 versus 2.154), matching the safety rejection. CogVideoX 0.15 shows the intended rescue: the selected path improves PSNR from 18.196 to 24.202, LPIPS from 0.918 to 0.344, OffRel from 5.322 to 0.338, and five-cycle drift from 4.841 to 0.662. WAN gives the damping case: at center 0.05, the selected path reaches PSNR 18.100 and LPIPS 0.618 while reducing OffRel and five-cycle drift to 0.310 and 0.690, whereas full CM remains closer to the round-trip boundary.
WAN 0.25 provides a second damping example. Relative to the latent filter, the selected path improves PSNR by $3.04$ dB while reducing OffRel from $0.709$ to $0.608$ and five-cycle drift from $1.371$ to $1.204$. Full CM achieves slightly lower LPIPS ($0.618$ versus $0.660$) but worsens both stability measures, so interpolation preserves most of its fidelity gain without inheriting its drift.

The full-test diagnostic sweep covers 18 representative emitted cells---five each for WAN, CogVideoX, and HunyuanVideo and three for Open-Sora---and all 18 pass LFV on 128 held-out clips (Table~\ref{tab:diag}). LPIPS, temporal-difference error, temporal high-frequency error, and repeated-cycle drift provide complementary decoded and temporal corroboration of the capacity-limited, damping-sensitive, and frontier regimes identified by LFV.

\begin{table}[t]
\centering
\caption{Full-test diagnostics for representative emitted cells. Margins are the weakest within each model group.}
\label{tab:diag}
\small
\begin{adjustbox}{max width=\linewidth}
\begin{tabular}{lrrrr}
\toprule
Model & Cells & Min PSNR LB & Max OffRel UB & Pass \\
\midrule
WAN & 5 & +1.202 & -0.043 & 5/5 \\
CogVideoX & 5 & +1.783 & -1.075 & 5/5 \\
Open-Sora v1.3 & 3 & +4.729 & -0.120 & 3/3 \\
HunyuanVideo & 5 & +2.724 & -0.017 & 5/5 \\
\midrule
Total & 18 & -- & -- & \textbf{18/18} \\
\bottomrule
\end{tabular}
\end{adjustbox}
\end{table}
\begin{table}[t]
\centering
\caption{Mean online latency (ms) at center 0.25 on an NVIDIA L40S, FP16, batch size 1.}
\label{tab:runtime}
\small
\begin{adjustbox}{max width=\linewidth}
\begin{tabular}{lrrrr}
\toprule
Model & Latent filter & Selected path & Pixel reencode & Speedup \\
\midrule
WAN & 105.7 & 105.5 & 334.9 & $3.17\times$ \\
Open-Sora v1.3 & 178.6 & 178.4 & 507.1 & $2.84\times$ \\
CogVideoX & 245.2 & 244.8 & 700.7 & $2.86\times$ \\
HunyuanVideo & 190.0 & 189.6 & 616.4 & $3.25\times$ \\
\midrule
Average & \textbf{179.9} & \textbf{179.6} & \textbf{539.8} & $\mathbf{3.01\times}$ \\
\bottomrule
\end{tabular}
\end{adjustbox}
\end{table}

\subsection{Additional Validation}
Three checks reinforce the operating-map interpretation. Replacing the decode-target with source-filter-reencode preserves all 20 mechanism-slice decisions under frozen selections. No-op passes only a small number of cells, while every emitted path has a strictly positive paired fidelity lower bound relative to no-op. Decode--encode projection reaches Open-Sora centers 0.075--0.75, including the latent-path frontier, demonstrating that the target remains reachable through a more expensive branch. Reconstruction attenuation alone also fails to predict the map: CogVideoX suppresses high-frequency reconstruction power more strongly than Open-Sora yet remains deployable across the radial grid. The decisive quantity is the learned round-trip response, not decoder bandwidth in isolation.

\subsection{Latent-Level Online Cost}
At inference, LFV reduces to a fixed per-frequency matrix response. Table~\ref{tab:runtime} shows that this response stays in the direct-latent cost regime across all four VAEs: average latency is 179.6 ms, essentially identical to the latent filter at 179.9 ms and $3.01\times$ faster than pixel filter--reencode at 539.8 ms. The method therefore converts offline measurement into online speed without inserting an additional VAE round trip.

Within each VAE, no-op, latent filter, C1, full CM, and selected path differ by less than 0.6 ms, so the measured cross-channel operator overhead is negligible relative to the shared output decode. The larger cross-model spread (105.5--244.8 ms) is therefore dominated by the model-specific VAE path; the accompanying peak-memory range, from about 0.94 GB for WAN to 8.07 GB for CogVideoX, is consistent with this interpretation. A selected response stores 4.001 MiB per VAE/edit setting, and the complete bank for the 423 emitted cells occupies about 1.65 GiB. This yields a practical deployment pattern: fit and validate a VAE-specific operator bank once, query the operating map for the requested spectral edit, and execute the selected response at latent-filter speed.

\section{Conclusion}
Video VAEs exhibit structured, model-specific spectral transfer rather than a universal correspondence between pixel and latent frequency. LFV measures this transfer, selects the amount of channel mixing supported by each VAE--edit pair, and compiles the result into a fixed latent operator. Across 544 cells, LFV emits 423 operators: 277 are diagonal-sufficient and 146 require cross-channel response. Thus, more than one third of the emitted operator bank depends on channel coupling rather than independent latent-channel gains.

The operating maps expose distinct model signatures and generalize across radial band-pass, low-pass, high-pass, notch, spatial, and temporal operators. Fully frozen OpenVid-fitted responses also pass all 20 generated-domain evaluations on CogVideoX and HunyuanVideo, demonstrating transfer beyond reconstruction clips without domain-specific adaptation. At inference, the selected response matches direct latent-filter latency and is about $3\times$ faster than pixel filter--reencode. LFV therefore converts offline VAE-specific measurement into a practical operator bank for fast, high-fidelity spectral control.

\bibliography{aaai2027}

\end{document}